\documentclass[journal]{IEEEtran}
\usepackage{amsthm}
\usepackage{algorithm}
\usepackage{algorithmic}
\usepackage[switch]{lineno}
\usepackage{xcolor} 
\usepackage{amsmath,amssymb,amsfonts}
\usepackage{graphicx}
\usepackage{tabularx}
\usepackage{booktabs}
\usepackage{float} 
\usepackage{comment}
\usepackage[hidelinks]{hyperref}
\usepackage[numbers]{natbib}

\usepackage{multirow}
\usepackage{nicematrix}

\begin{document}

\markboth{This paper is accepted in the \textbf{IEEE Transactions on Artificial Intelligence (TAI)}}%
	    {This paper is accepted in the \textbf{IEEE Transactions on Artificial Intelligence (TAI)}}

\title{Label-efficient Time Series Representation Learning: A Review}

\author{Emadeldeen Eldele, Mohamed Ragab, Zhenghua Chen, Min Wu, Chee-Keong Kwoh \\ and Xiaoli Li,~\IEEEmembership{IEEE~Fellow }

\thanks{This work is supported by the Agency for Science, Technology and Research (A$^*$STAR) Singapore under its NRF AME Young Individual Research Grant (Grant No. A2084c1067).}
\thanks{Emadeldeen Eldele and Xiaoli Li are with the Institute for Infocomm Research, Agency for Science, Technology and Research, Singapore, Centre for Frontier AI Research, Agency for Science, Technology and Research, Singapore, and also with the School of Computer Science and Engineering at Nanyang Technological University, Singapore (E-mails: emad0002@ntu.edu.sg and xlli@i2r.a-star.edu.sg).}
\thanks{Mohamed Ragab is with the Institute for Infocomm Research, Agency for Science, Technology and Research, Singapore, Centre for Frontier AI Research, Agency for Science, Technology and Research, Singapore (E-mail: mohamedr002@e.ntu.edu.sg).}
\thanks{Zhenghua Chen is with the Institute for Infocomm Research, Agency for Science, Technology and Research, Singapore and the Centre for Frontier AI Research, Agency for Science, Technology and Research, Singapore (E-mail: chen0832@e.ntu.edu.sg).}
\thanks{Min Wu is with the Institute for Infocomm Research, Agency for Science, Technology and Research, Singapore (E-mail: wumin@i2r.a-star.edu.sg).}
\thanks{Chee-Keong Kwoh is with the School of Computer Science and Engineering, Nanyang Technological University, Singapore (E-mail: asckkwoh@ntu.edu.sg).}
\thanks{Min Wu is the corresponding author.}}

\maketitle

\begin{abstract}
Label-efficient time series representation learning, which aims to learn effective representations with limited labeled data, is crucial for deploying deep learning models in real-world applications. To address the scarcity of labeled time series data, various strategies, e.g., transfer learning, self-supervised learning, and semi-supervised learning, have been developed. In this survey, we introduce a novel taxonomy for the first time, categorizing existing approaches as in-domain or cross-domain, based on their reliance on external data sources or not. Furthermore, we present a review of the recent advances in each strategy, conclude the limitations of current methodologies, and suggest future research directions that promise further improvements in the field.
\end{abstract}

\begin{impact}
This research demystifies the application of deep learning to time series data with limited labels. It sets the stage for significant technological advancements, enabling improved model performance in healthcare, finance, and environmental monitoring without the extensive need for labeled data. Economically, it offers a pathway to cost savings in data processing, while socially, it could improve public health and environmental management. By simplifying complex AI methodologies for non-specialists, this work not only progresses the AI field but promises widespread benefits across various sectors, making deep learning more accessible and impactful in real-world applications.
\end{impact}


\begin{figure*}
    \centering
    \includegraphics[width=0.9\textwidth]{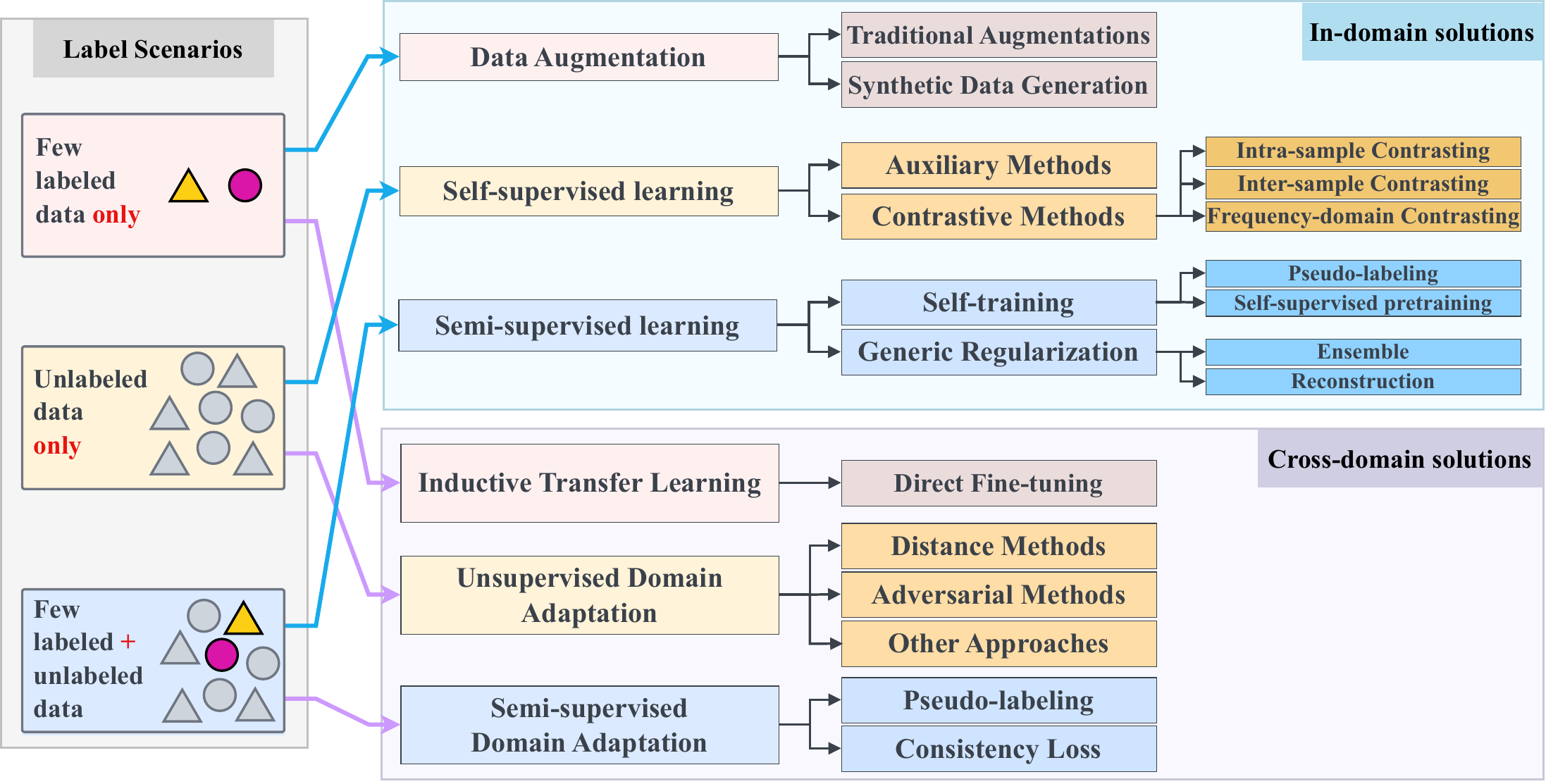}
    \caption{Graphical overview of the different scenarios of data availability and the suitable solution. Note that the directions given by arrows are not the only solutions for each respective scenario, but can be the best way to achieve the best performance.}
    \label{fig:few_lbls_scenarios}
\end{figure*}

\section{Introduction}\label{sec:introduction}
Deep learning has shown significant success in diverse time series applications. Unfortunately, the applicability of deep learning models in real-world scenarios is tied to the availability of \textit{large} and \textit{labeled} datasets, which are usually lacking for time series applications \cite{RASHID2019100944,fawaz2019deep}. For example, we can collect massive amounts of sensory data from manufacturing sensors, while only a few of them are being annotated by domain experts.
Also, in some healthcare applications, it can be difficult to collect or access enough data, due to security reasons or the patient’s right to privacy, where we end up with only a few (labeled) samples \cite{abouelmehdi2018big}. In either situation, training efficient deep learning models with these data and annotation limitations can be very challenging.

Over recent years, a significant body of research has emerged to address the label scarcity challenges, with methodologies spanning various conceptual paradigms \cite{ma2023survey,da_ts_survey}. These methodologies can be categorized into in-domain methods and cross-domain methods.
In-domain methods focus on deriving representations from the available time series data, without the reliance on external data sources. Strategies within this category include data augmentation, which increases the number of labeled samples. Additionally, self-supervised learning allows learning from fully unlabeled data \cite{simclr}, while semi-supervised learning provides a harmonized approach to learning from both labeled and unlabeled instances within the same domain \cite{fixmatch}. Conversely, cross-domain methods learn representations from an external, labeled source domain, subsequently transferring this acquired knowledge to the target domain of interest. The common strategy is to use transfer learning approaches, either by directly fine-tuning the pretrained model on the target domain or by adapting the source and target domains to have similar distributions and class boundaries \cite{da_ts_survey}.

In this survey, we navigate through three specific real-world scenarios characterized by the availability of data: scenarios with a limited number of labeled samples, those with exclusively unlabeled time series data, and situations combining both labeled and unlabeled data. We introduce a novel taxonomy that systematically organizes the solutions to these scenarios into in-domain and cross-domain strategies. This taxonomy, depicted in Fig.~\ref{fig:few_lbls_scenarios}, guides the classification of each solution based on its reliance on external data sources.
Our objective extends beyond merely cataloging recent advancements within these categories. We aim to craft a cohesive roadmap that guides the effective deployment of deep learning models despite the constraints posed by limited labeled data. It is important to note that our intention \textit{is not} to provide an exhaustive discussion of each method within every category. Instead, we strive to offer a panoramic view, presenting a high-level understanding of the challenges and available solutions. By doing so, we enable readers to grasp the overarching concepts and approaches, facilitating further exploration and in-depth investigations according to their specific research interests.

Overall, the contributions of this paper can be summarized as follows:
\begin{itemize}
    \item We conduct a comprehensive survey of label-efficient learning techniques tailored for time series data. This includes scenarios with limited labeled samples, entirely unlabeled data, and their combinations, providing a detailed overview of the current landscape and the inherent challenges of limited labeled data.

    \item We propose a novel taxonomy that systematically categorizes existing approaches into in-domain and cross-domain strategies, offering a new lens through which to understand and navigate the solutions addressing label scarcity in time series analysis. It underscores the dependency of these strategies on external data sources, presenting a unique framework for future research and application.

    \item We explore the challenges, emerging trends, and future directions in label-efficient time series analysis. 
\end{itemize}

\section{Related Surveys}
In the realm of label-efficient time series representation learning, several surveys have delved into specific solutions tailored to address the challenges of limited labeled data \cite{ts_aug_survey,ssl_survey_ts,zhang2023selfsupervised,ma2023survey,fawaz_tl,da_ts_survey}. While these surveys have significantly contributed to the field, they tend to focus on specific techniques or solutions, limiting their scope to a singular aspect of label-efficient time series representation learning. Our survey, on the other hand, synthesizes these individual perspectives into a cohesive narrative that encompasses a broader range of strategies. 

We first discuss surveys related to in-domain solutions. \citet{ts_aug_survey} comprehensively summarize time series data augmentation techniques across key tasks such as forecasting, anomaly detection, and classification. The authors systematically review different augmentations from simple time-domain transformations to more intricate methods in transformed frequency and time-frequency domains.
Also, \citet{ssl_survey_ts} explore the self-supervised representation learning to learn from unlabeled time series data across various modalities, with an emphasis on temporal data and cross-modal learning models. The authors present a classification scheme to benchmark, and categorize various studies, as a guide for model selection.

Furthermore, \citet{zhang2023selfsupervised} present an exhaustive review of state-of-the-art self-supervised learning techniques specifically tailored for time series data. They categorize these methods into three distinct paradigms: generative-based, contrastive-based, and adversarial-based, and they delve into various frameworks and techniques within each category.
Similarly, \citet{ma2023survey} thoroughly examine time series pre-trained models, where they discuss time series modeling tasks and the associated deep learning models. They also discuss supervised, unsupervised, and self-supervised pre-training techniques.

Regarding the cross-domain related surveys, inspired by the success of transfer learning in computer vision, \citet{fawaz_tl} explored the potential of transfer learning for time series classification and conducted exhaustive experiments to draw conclusions. Their findings underscore the presence of transferable low-level features in time series, and their transferability can be measured by the Dynamic Time Warping technique. Last, \citet{da_ts_survey} offer a thorough discussion about unsupervised domain adaptation in the context of time series sensor data. They categorize these studies by industry, detailing application backgrounds, frequently used sensors, data discrepancy sources, and prominent research datasets.

Concluding our discussion on the related surveys, we emphasize that, to our knowledge, our work is the first to comprehensively address label-efficient representation learning for time series data from various perspectives. We provide a broad overview, exploring multiple scenarios and their solutions in depth.

\section{Preliminaries}
\subsection{Literature Search and Selection Strategy}
Our review of time series representation learning for label-efficient scenarios was founded on a systematic search and selection methodology, tailored to capture the recent advances and the relevant contributions to the field.

Search Strategy: We utilized major academic databases, including IEEE Xplore, PubMed, Google Scholar, and ACM Digital Library, focusing on publications from the past five years. Our search was guided by carefully chosen keywords including ``time series analysis", ``deep learning", ``self-supervised learning", ``semi-supervised learning", ``unsupervised domain adaptation", ``transfer learning", ``semi-supervised domain adaptation", ``weakly supervised learning", ``active learning", and ``prototypical networks", each appended with ``time series" or some related applications, e.g., ``EEG", ``Human Activity Recognition", and ``fault diagnosis", to focus specifically on time series contexts.

\emph{Selection Criteria:} Articles were chosen based on peer-review status and direct relevance to label-efficient learning in time series analysis. We excluded non-peer-reviewed sources, except for those showing high impact on the field, e.g., \cite{bai2018empirical,cpc} or recent surveys. We also excluded unrelated topics, and older publications, unless they offered foundational insights.

\emph{Selection Process:} An initial screening of titles and abstracts was followed by a full-text review to assess each article's contribution to the field. This dual-phase process ensured a comprehensive and unbiased selection of literature.
This streamlined approach enabled us to distill significant advancements and methodologies in time series representation learning, addressing the challenges of label scarcity.

\subsection{Representation Learning Problem Formulation}
In this section, we present a mathematical framework that encapsulates the essence of time series label-efficient learning, covering both in-domain and cross-domain learning scenarios.

Consider a time series dataset \( \mathcal{D} \) composed of pairs \( (\mathbf{x}, \mathbf{y}) \), where \( \mathbf{x} = \{x_1, x_2, ..., x_N\} \) represents the collection of time series data points and \( \mathbf{y} = \{y_1, y_2, ..., y_N\} \) denotes their corresponding labels. In label-efficient learning scenarios, the goal is to maximize the predictive performance of a learning model \( f: \mathcal{X} \rightarrow \mathcal{Y} \) while minimizing the dependence on labeled instances. Here, \( \mathcal{X} \) and \( \mathcal{Y} \) represent the input and output spaces of all possible time series instances, respectively. Each time series instance \( x_i \in \mathbf{x} \) is a sequence of observations \( x_i = \{o_1, o_2, ..., o_T\} \), where \( T \) is the length of the series and \( o_t \) represents the observation at time \( t \). For supervised scenarios, each \( x_i \) has a corresponding label \( y_i \in \mathbf{y} \). In label-efficient scenarios, \( \mathbf{y} \) may be partially observed or observed with noise.

\textbf{In-Domain Learning:} The dataset \( \mathcal{D} \) can vary in composition, encompassing labeled samples only \( \mathcal{D}_L = \{(x_i, y_i)\}_{i=1}^{L} \) if \( L = N \), unlabeled samples only \( \mathcal{D}_U = \{x_j\}_{j=1}^{E} \) if \( E = N \), or a combination of both subsets if \( L + E = N \). The learning objective adapts to this variability, aiming to utilize \( \mathcal{D}_L \), \( \mathcal{D}_U \), or their integration to effectively train a model \( f \) that can accurately predict \( \mathbf{y} \) for new instances, accommodating the dataset's specific labeling scenario.

\textbf{Cross-Domain Learning:} Here, we consider two datasets: a source domain \( \mathcal{D}_S = \{(x^S_i, y^S_i)\} \) with abundant labels and a target domain \( \mathcal{D}_T = \{(x^T_j, y^T_j)\}_{j=1}^{M} \) where labels may be scarce or absent. The objective is to adapt the knowledge from \( \mathcal{D}_S \) to improve learning on \( \mathcal{D}_T \).

\subsection{Deep Learning Architectures}
\label{sec:arch}
Selecting an appropriate architecture is foundational to the construction of a label-efficient deep learning model. Various deep learning architectures have been adopted, summarized as follows.
\subsubsection{Convolutional Neural Networks (CNNs)}
CNNs, including 1D-CNN \cite{7966039}, 1D-ResNet \cite{resnet_reference}, and temporal convolutional network (TCN) \cite{bai2018empirical}, are prevalent in time series classification. Other designs have been proposed, introducing different designs, e.g., InceptionTime \cite{ismail2020inceptiontime} and Omni-Scale CNN blocks \cite{tang2021omni}. These networks excel at identifying patterns and trends with fewer parameters, reducing overfitting risks.

\subsubsection{Recurrent Neural Networks (RNNs)}
RNNs, including LSTM models, are tailored for time series data, incorporating memory components to leverage past timesteps \cite{NEURIPS2018_5cf68969}. They usually excel in forecasting and can be utilized alongside or following CNNs in classification tasks \cite{khan2021bidirectional,deepsleepnet}. While effective in learning from historical data, RNNs struggle with long sequences and are computationally intensive due to their sequential nature.

\subsubsection{Attention Mechanism}
The attention mechanism was proposed to enhance LSTM by focusing on specific input parts and assigning scores to hidden states \cite{LI2019104785}. Later, it was adopted by several time series applications \cite{LIU2022103001,dsanet,attnSleep}. Recently, the Transformer model became emerging as a prominent architecture for time series \cite{zerveas2021transformer,ecgTransform}. While attention improves long sequence handling and model interpretability, its computational cost and overfitting risk due to more parameters are notable drawbacks.

For the following sections, the block ``Encoder'' in diagrams is generic and can be adopted as one of the above networks, allowing for flexibility depending on the specific requirements and constraints of the application.

\section{In-domain Representation Learning}
In this section, we discuss three approaches to learning representations from the data in hand. First, we assume possessing only a few labeled samples which can be addressed with data augmentation.
Second, we assume the existence of a large cohort of unlabeled data and we discuss how self-supervised learning can be utilized.
Third, we may have some labeled samples besides the massive unlabeled data, where we can apply the semi-supervised learning techniques.

\subsection{Data Augmentation}
\label{sec:augmentaionts}
\begin{figure}
    \centering
    \includegraphics[width=\columnwidth]{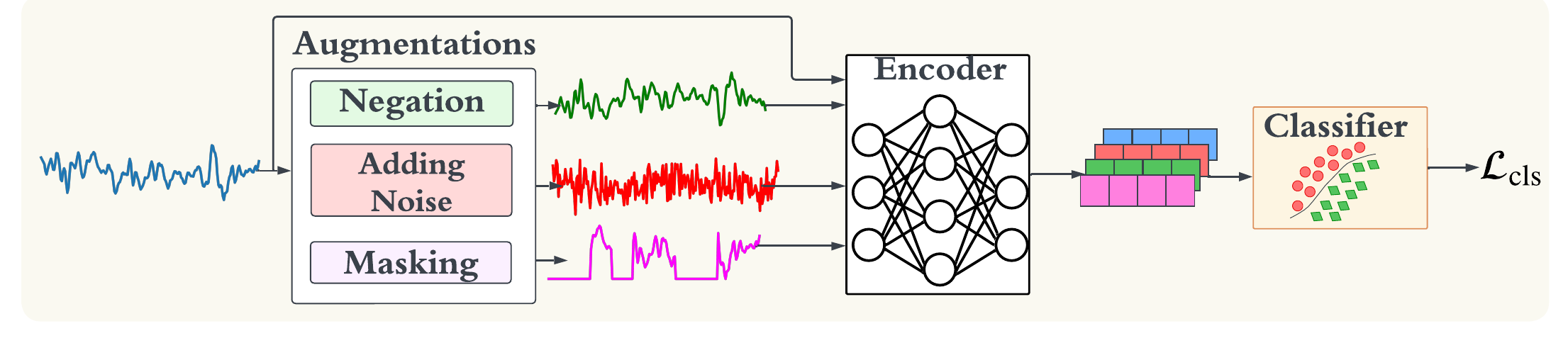}
    \caption{Data augmentation is utilized to increase the number of training samples. For each input sample, several augmentations are applied to generate new samples.}
    \label{fig:augmentations}
\end{figure}

Data augmentation is usually adopted as a way to artificially scale up the training dataset, by applying several transformations to the input data. This can be particularly useful when it is difficult or expensive to collect more real-world data. Fig.~\ref{fig:augmentations} illustrates the principle of data augmentation applied to a single time series example. The different augmentations shown in the figure elucidate how one original sequence can be transformed into multiple variants, each serving as a distinct training example.

Given a labeled dataset \( \mathcal{D}_L = \{(x_i, y_i)\}_{i=1}^{L} \), data augmentation aims to generate an augmented dataset \( \mathcal{D}_{\text{aug}} = \{(\tilde{x}_i, \tilde{y}_i)\} \) where \( \tilde{x}_i \) is a modified version of \( x_i \) obtained through various augmentation techniques (e.g., warping, noise injection), and \( \tilde{y}_i = y_i \) assuming the label is invariant to the augmentation. The objective is to train a model \( f \) to improve generalization, and minimize \( \mathcal{L}(f(\tilde{x}_i), \tilde{y}_i) \) over \( \mathcal{D}_{\text{aug}} \cup \mathcal{D}_L \).

By adding more samples, data augmentation aims to provide the model with diverse instances that are representative of the underlying data distribution. These transformed samples not only contribute to better generalization but also alleviate the need for an excessive volume of labeled data. Aside from increasing the sample size, the variety of augmentations applied to the time series data can generate new sequences that are realistic variations of the original data, mimicking natural fluctuations and nuances. Thus, the model can become robust against different orientations, scales, noise, and other variations that might be present in real-world data. In addition, some works utilize data augmentation to upsample the minority classes to overcome the class imbalance issue in some time series applications \cite{Fan_2020}.

There are several data augmentation techniques proposed for time series data. \citet{ts_aug_survey} provide a comprehensive set of time series data augmentations that can be applied in the time domain and the frequency domain. They categorized the augmentations into basic and advanced approaches and provide several subcategories under each, e.g., decomposition
methods and statistical generative models. Similarly, \citet{iwana2021empirical} conduct a survey of data augmentation for empirically evaluating the classification tasks, where they explore data transformations and the different ways to mix time series patterns. In general, some augmentations are more widely employed, e.g., scaling, shifting, rotation, masking, and noise injection.

Besides applying transformations, synthetic data generation is yet another technique to artificially generate data using simulation. This technique can be particularly useful in preserving data privacy, as it might not be viable to access real-world data in some cases. Therefore, generating synthetic data provides a way to use the data without compromising privacy. There are several different methods to generate synthetic data, but we mainly explore the Generative Adversarial Networks (GANs), being the most commonly used approach. For example, ActivityGAN generates time series sensory data for human activity recognition with 1D-CNN networks by training a GAN \cite{ActivityGAN}. They find that performance increases by adding more synthetic data to real data.
Similarly, DCGAN generates artificial EEG signals to improve the performance of the Brain-Computer Interface (BCI) tasks \cite{eeg_gan}, where the generator receives the subject-specific feature vector to generate artificial samples. By training it against the discriminator, it learns to generate artificial samples similar to the original ones. Their model witnessed a significant improvement by including the generated data.

\subsection{Self-supervised Learning}
\begin{figure}
    \centering
    \includegraphics[width=\columnwidth]{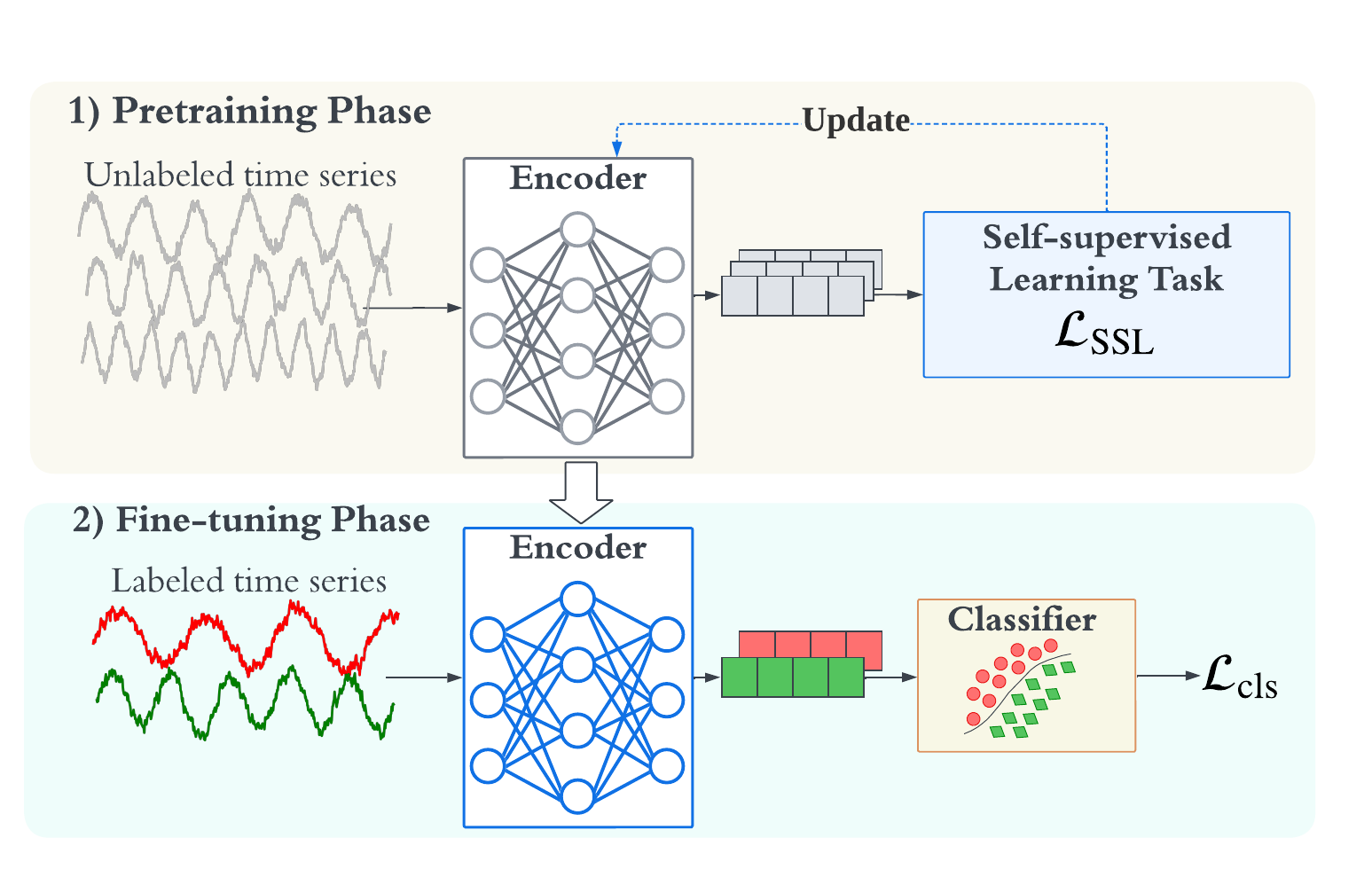}
    \caption{The pipeline of the self-supervised learning process. In the pretraining phase, the encoder is trained with the self-supervised loss without labels. Next, in the fine-tuning phase, the pretrained encoder is fine-tuned with the available few labeled data.}
    \label{fig:self_supervised_learning}
\end{figure}

Self-supervised learning involves training a model to learn representations from data using only the unlabeled samples. Instead of relying on human-provided labels, we define a new task to be solved by the model or generate pseudo labels based on the input data. The general process of self-supervised learning is shown in Fig.~\ref{fig:self_supervised_learning}. The training process is usually performed in two phases. In the first phase, i.e., the pre-training phase, we feed the encoder model with the unlabeled data and train it with the self-supervised task. Next, we use the few labeled samples in the second phase, i.e., the fine-tuning, to tailor the model to the downstream task, and improve its performance on these few labels, compared to fully supervised training with the few labels.

Self-supervised learning methods can be categorized into auxiliary- or contrastive-based methods. In both categories, the unlabeled data is used to learn an additional task that is not the main focus of the model but can provide additional supervision for learning useful representations of the data. 

\subsubsection{Auxiliary-based Methods}
The auxiliary-based methods propose an auxiliary task that can be related to the downstream task or the characteristics of the data, which helps the pretrained model to perform well on the downstream task. Given an unlabeled dataset \( \mathcal{D}_U = \{x_j\}_{j=L+1}^{N} \), we define a pretext task that generates pseudo-labels \( \hat{y}_j \) for \( x_j \). The model \( f \) learns to predict \( \hat{y}_j \) from \( x_j \), effectively learning useful representations without manual labeling.

For example, \citet{banville2020uncovering} introduce two auxiliary tasks, i.e., relative positioning and temporal shuffling, to learn the underlying temporal characteristics in the EEG signals. In the first task, a window of the signal is used as an anchor to measure the relative position of other windows, while in the second task, a window of the signal is shuffled to teach the model to fix the temporal order of the signal. Similarly, \citet{sense_learn} propose some auxiliary tasks, e.g., feature prediction from a masked window, temporal shift prediction, modality denoising, and blend detection, to learn representations from human activity recognition data.
In addition, TST \cite{zerveas2021transformer} and STraTS \cite{STraTS} propose to predict the masked part in the signal as an auxiliary task to pre-train a Transformer model. Furthermore, \citet{islam2023personalized} introduce a personalized stress prediction system that utilizes wearable multimodal biosignal time series data. Their approach relies on a self-supervised forecasting task for pre-training models on individual subjects, then fine-tuning specifically for the task of stress prediction.

Another example of auxiliary tasks is to utilize data augmentations to provide several transformations to the time series data. Each transformation is assigned a pseudo label, and the model is trained to classify them with a standard cross-entropy loss. Several works follow this approach in different applications, e.g., electrocardiogram (ECG) emotion recognition \cite{ecg_ssl} and human activity recognition \cite{har_ssl}.

Even though auxiliary tasks have shown effectiveness, they restrict the generality of the learned representations \cite{cpc}. Therefore, most time series self-supervised learning works have recently adopted contrastive learning instead \cite{emadeldeen2023eval}.

\subsubsection{Contrastive-based Methods}
Contrastive-based self-supervised learning methods aim to form positive and negative pairs, where the contrastive loss maximizes the similarity among positive pairs while minimizing the similarity among negative pairs. 
To illustrate, given an unlabeled dataset \( \mathcal{D}_U = \{x_j\}_{j=1}^{N} \) of time series instances, the objective of contrastive self-supervised learning is to learn a representation \( f \) that embeds similar instances, i.e., positive pairs closer together and dissimilar instances, i.e., negative pairs, further apart in the embedding space. The positive pairs are data instances \( (x_i, x_i^+) \) derived from the same time series instance or through augmentations that preserve the semantic content (see Section~\ref{sec:augmentaionts}). The negative pairs are \( (x_i, x_j^-) \), where \( x_i \) and \( x_j^- \) do not share the same semantic content, often chosen from different time series instances within the dataset. The objective is to learn a function \( f \) that minimizes a contrastive loss, such as the InfoNCE loss \cite{cpc}, defined for a batch of instances as follows:
\begin{equation}
\mathcal{L}_{\text{contrastive}} = -\sum_{i=1}^{n} \log \frac{\exp(sim(f(x_i), f(x_i^+)) / \tau)}{\sum_{x_j^- \in N} \exp(sim(f(x_i), f(x_j^-)) / \tau)},
\end{equation}
where \( sim(u, v) \) is a similarity measure between two embeddings (e.g., the dot product or cosine similarity), \( \tau \) is a temperature parameter that scales the similarity measure. Notably, the sum in the denominator runs over a set of negative pairs for each positive pair, and \( n \) is the number of positive pairs in a batch. In addition, the hyperparameter \( \tau \) plays a crucial role in controlling the scale of the distribution of distances between embeddings. A smaller value of \( \tau \) sharpens the distribution increasing the sensitivity to small differences in distances, potentially focusing the learning on hard negatives. Conversely, a larger \( \tau \) value smooths out the distribution, which can mitigate the impact of outliers but may reduce discriminative power by treating dissimilar pairs more leniently.

Previous works propose different strategies to choose the positive and negative pairs to improve the quality of the learned representations. Additionally, most works attempt to learn the temporal relations within time series in the pretraining phase. In this section, we provide a new time series-specific categorization for contrastive-based self-supervised learning methods. Particularly, we categorize them into intra-sample contrastive methods, inter-sample contrastive methods, and frequency domain contrastive methods.

\paragraph{Inter-sample Contrasting (Contextual Contrasting)}
In inter-sample contrastive learning, the model uses the input signal as a whole without splitting or windowing. One of the well-performing visual contrastive methods is SimCLR \cite{simclr}, which relies on data augmentations to define positive and negative pairs. In specific, the positive pairs are the augmented views of an anchor sample, while the negative pairs are all the augmented views of other samples. Many works followed SimCLR for time series data, such as \cite{simclr_SSC1,POPPELBAUM2022_FDsimclr} that apply SimCLR-like methods for sleep stage classification and fault diagnosis tasks respectively.
In addition, TimeCLR~\cite{TimeCLR} adapts the SimCLR framework besides introducing DTW-based data augmentation and utilizing InceptionTime for feature extraction to tackle the challenge of sparse labels.

ExpCLR~\cite{expCLR} leverages expert knowledge instead of relying on data transformations for contrastive learning. Recognizing the availability of expert features in fields like industry and medicine, this method incorporates such features to ensure the development of useful time-series representations that fulfill two proposed essential properties often overlooked by current methods, i.e., the proximity of representations for samples with similar expert features and the separation of representations for samples with markedly different expert features. These properties are essential in defining a new loss function tailored to handle continuous expert features.
Mixup Contrastive Learning (MCL) proposes to replace the traditional contrastive loss with a mixup-based contrastive loss \cite{WICKSTROM2022mixupContrastive}. Specifically, MCL generates new mixed samples in each mini-batch using the mixup operation, and the proposed loss predicts the mixup ratio among each pair of samples. 
Also, \citet{triplet_loss} propose another loss that searches for positive and negative pairs for each anchor sample in both the input and embedding spaces. The contrastive loss is then applied to the selected triplet to learn a causal temporal CNN.

MHCCL~\cite{MHCCL} learns semantic-rich representations from unlabeled multivariate time series data, utilizing a Masked Hierarchical Cluster-wise Contrastive Learning framework. It addresses the issue of false negative pairs in traditional contrastive learning by exploiting hierarchical clustering for semantic information at multiple granularities. With downward and upward masking strategies, MHCCL efficiently filters out fake negatives and refines cluster prototypes, enhancing the clustering process and quality. 
The idea of supervised contrastive loss was also adopted by MICOS~\cite{MICOS}, where they blend self-supervised, intra-class, and inter-class supervised contrastive learning to effectively utilize available labels to address the challenges in capturing complex spatio-temporal features inherent in multivariate time series. Last, InfoTS proposes a contrastive learning framework that utilizes information theory to guide the selection of data augmentations, ensuring both high fidelity to the original data and sufficient variety to foster robust and discriminative representation learning \cite{10.1609/aaai.v37i4.25575}. By applying information theory, InfoTS quantitatively evaluates potential augmentations based on their ability to preserve crucial information content of the time series data, thereby ensuring the augmented samples are representative and beneficial for learning discriminative features.

\paragraph{Intra-sample Contrasting (Temporal Contrasting)}
Intra-sample contrastive methods aim to learn the temporal relations in time series data in the pretraining phase. To do so, they process the signal at the timestep level, either in the embedding space or in the input space. One of the earliest works is Contrastive Predictive Coding (CPC), which learns the temporal relations by splitting the embedding space of the signal into a past part and a future part \cite{cpc}. Then, an autoregressive model is trained to produce a context vector from the past part and use it to predict the future part using contrastive learning. Similarly, SleepDPC generates multiple context vectors and uses each one to predict the next timestep in the future part for the sleep stage classification task \cite{9414752}.
Also, TS-TCC applies strong and weak augmentations for each input signal, generates the embeddings for each, and uses the past part of each view to predict the future part of the other view in a tough cross-view prediction task \cite{tstcc}. Last, ContrNP replaces the autoregressive model used by CPC with a neural process for time series forecasting \cite{neural_process}, effectively bypassing the need for manually designed data augmentations. By leveraging a variety of sampling functions, this approach generates augmented data variants, extending neural processes with a novel contrastive loss to learn time series representations in a self-supervised manner.

Other methods learn the temporal relations among timesteps in the input space. For example, Temporal Neighborhood Coding (TNC) assumes that the distribution of one window of timesteps is similar to its neighboring windows and can be more distinguishable than the non-neighboring windows \cite{tonekaboni2021unsupervised}. Therefore, TNC forms positive pairs among neighboring windows and negative pairs among non-neighboring windows. Another example is TS2Vec, which applies data augmentation on the input signal and considers the current timestep from the augmented views as positives, while the different timesteps from the same time series are considered as negatives \cite{yue2022ts2vec}. TS2Vec utilizes a hierarchical contrastive learning approach across augmented context views for robust contextual timestamp representations, which allows for simple aggregation to represent any sub-sequence within a time series.

\paragraph{Frequency Domain Contrastive Methods}
A recent trend in time series contrastive learning is to utilize the frequency domain characteristics in representation learning, being one main distinguishing properties of time series data. For instance, Bilinear Temporal-Spectral Fusion (BTSF) applies an iterative bilinear fusion between feature embeddings of both time and frequency representations of time series \cite{icml2022iterative}. Specifically, BTSF generates different views of time series and applies iterative aggregation modules to perform cyclic refinement to the temporal and spectral features. Similarly, both STF-CSL \cite{STFNets} and TF-C \cite{zhang2022_tfConsistency} apply time domain and frequency domain augmentations, and push the time domain and frequency domain representations of the same sample closer to each other while pushing them apart from representations of other signals. In particular, STF-CSL \cite{STFNets} integrates Short-Time Fourier Neural Networks (STFNets) as the core component and applies both time-domain and frequency-domain data augmentations. STF-CSL aligns with the physical reality of IoT sensing scenarios while also offering substantial performance improvements over traditional time-domain-based self-supervised methods in tasks such as human activity recognition. TF-C~\cite{zhang2022_tfConsistency} introduces a pre-training model focused on overcoming the challenges associated with mismatches in temporal dynamics between pre-training and target domains in time series data. TF-C embeds time-based and frequency-based representations of an example closely together in the time-frequency space, without requiring access to target domain examples during pre-training.

\begin{figure}
    \centering
    \includegraphics[width=\columnwidth]{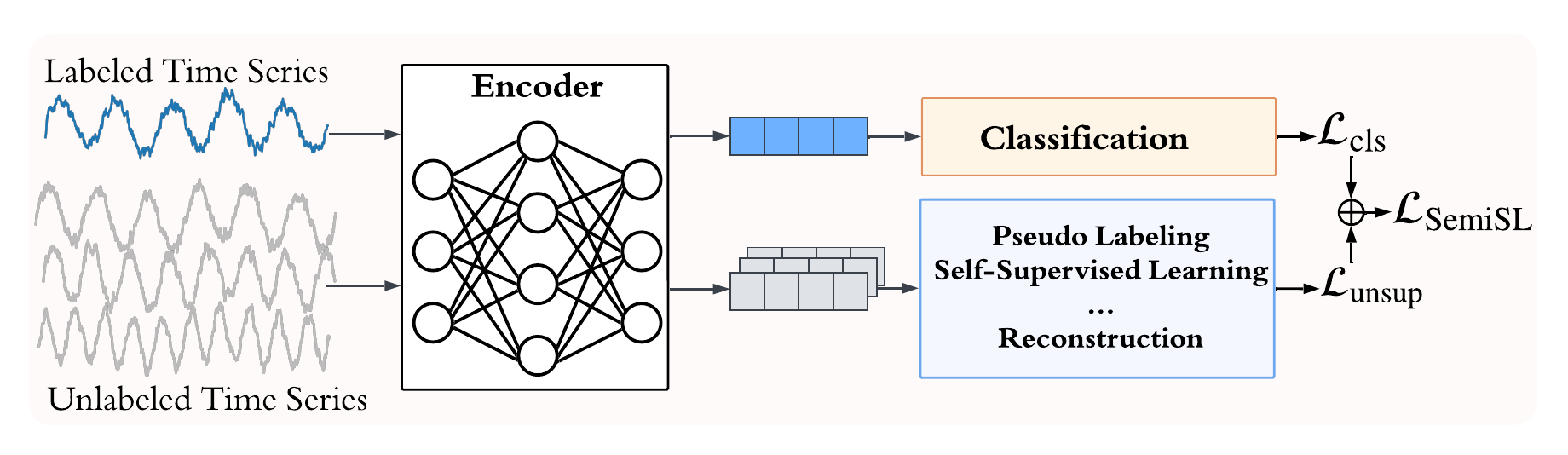}
    \caption{Illustration of the semi-supervised learning process. We train the model with the labeled portion of the data via the cross-entropy loss \(\mathcal{L}_{\text{cls}}\), and the unlabeled portion via an unsupervised learning technique \(\mathcal{L}_{\text{unsup}}\), i.e., \( \mathcal{L}_{\text{SemiSL}} = \mathcal{L}_{\text{cls}} +  \mathcal{L}_{\text{unsup}} \).}
    \label{fig:semi_supervised_learning}
\end{figure}

\subsection{Semi-supervised Learning}
Semi-supervised learning is a type of learning that involves training a model using both labeled and unlabeled data. 
It becomes beneficial when it is difficult or expensive to obtain a large amount of labeled data, and only a few labeled samples are available with a large amount of unlabeled data. 
Specifically, given a small labeled dataset \( \mathcal{D}_L \) with a larger unlabeled dataset \( \mathcal{D}_U \), the goal is to train a model \( f \) that leverages the unlabeled data \( \mathcal{D}_U \) alongside \( \mathcal{D}_L \) to enhance learning. The learning objective often combines a supervised classification loss \( \mathcal{L}_{\text{cls}}(f(x_i), y_i) \) on \( \mathcal{D}_L \) with an unsupervised loss \( \mathcal{L}_{\text{unsup}}(f(x_j)) \) on \( \mathcal{D}_U \), where \( \mathcal{L}_{\text{unsup}} \) encourages the model to learn meaningful structures from \( \mathcal{D}_U \). As illustrated in Fig.~\ref{fig:semi_supervised_learning}, the overall objective becomes \( \mathcal{L}_{\text{SemiSL}} = \mathcal{L}_{\text{cls}} +  \mathcal{L}_{\text{unsup}} \), where both losses are added together during training. However, one challenge in semi-supervised learning still achieving the best balance of learning from both labeled and unlabeled samples.

In this survey, we categorize semi-supervised learning approaches into self-training-based methods and generic regularization-based methods.

\subsubsection{Self-training Methods}
Self-training involves assigning labels to the unlabeled data to train the model. It can be implemented either by pseudo-labeling or self-supervised pretraining.

\paragraph{Pseudo-Labeling}
Pseudo labeling leverages a small set of labeled data to initially train a model, which is then used to generate pseudo labels for the unlabeled data, based on the assumption that these labels are likely accurate. The typical method for producing pseudo labels involves applying the softmax function to the model's predictions. Specifically, if a model predicts the output probabilities \(p_i\) for a data point, the pseudo label is assigned by selecting the class with the highest probability after applying softmax, which is mathematically represented as: $\text{pseudo-label} = \arg\max(\text{softmax}(p_i))$.

For example, \citet{sssl} generate pseudo labels from the model predictions to the unlabeled data and use least-square regression to learn both shapelets and classification boundaries. Another way is to use clustering techniques as in SUCCESS, which proposes a hierarchical clustering scheme to assign pseudo labels to samples according to their respective clusters \cite{success_semi}. It calculates the distance to centroids based on the dynamic time warping (DTW) distance metric.

A different way of utilizing the pseudo labels is proposed in FixMatch, which applies two augmentations to the image data, where one augmentation is weak and the other is strong \cite{fixmatch}.
Then, if the prediction of the weak-augmented view is beyond a threshold value, it uses its pseudo label for the strong augmented view. By showing success with image data, several methods applied FixMatch-like approaches on time series data, e.g., PARSE \cite{same_fixmatch} for EEG-based emotion recognition. PARSE employs pairwise representation alignment to bridge the gap between large volumes of unlabeled data and limited labeled data through a sequence of data augmentation, label guessing and sharpening, and representation alignment.
Also, SelfMatch uses signals with Gaussian noise as a weak augmentation and timecut as a strong augmentation \cite{selfMatch}. It also applies self-distillation to guide the lower-level blocks of the feature extractor with the knowledge obtained in the output layer.

\begin{table*}[htbp]
\centering
\caption{Summary of In-domain Methods for Time Series Representation Learning.}
\begin{NiceTabular}{c|c|c|p{1.7cm}|p{4.9cm}|p{4cm}|p{4cm}}
\toprule
\multicolumn{3}{c|}{\textbf{Category}} & \textbf{Method} & \textbf{Description} & \textbf{Advantages} & \textbf{Potential Limitations} \\

\midrule
\multirow[origin=c]{7}{*}{\rotatebox{90}{Data Augmentation}} & & & \cite{ts_aug_survey}, \cite{iwana2021empirical} & Applies transformations, e.g., scaling and masking to generate realistic variations of original time series data & Increases model robustness, alleviates need for labeled data & May introduce artificial bias if not carefully managed \\
\cmidrule{4-7}
&  & & ActivityGAN \cite{ActivityGAN} & Uses GAN to generate sensory data for human activity recognition & Helps in data privacy, improves performance with synthetic data & Quality of generated data heavily depends on GAN's training \\
\cmidrule{4-7}
&  & & DCGAN \cite{eeg_gan} & Generates artificial EEG signals to aid BCI tasks & Improves performance with the inclusion of generated data & Risk of generating non-representative or biased data \\
\hline
\midrule

\multirow[origin=c]{35}{*}{\rotatebox{90}{Self-supervised Learning}} & \multirow{10}{*}{\rotatebox{90}{Auxiliary-based Methods}} & & \cite{banville2020uncovering} & Introduces tasks to understand EEG signal's temporal dynamics & Enhances temporal understanding & May limit generality \\
\cmidrule{4-7}
& & & Sense and learn \cite{sense_learn} & Employs various tasks for representation learning in HAR & Diverse methods to handle data & Potentially restricts representation generality \\
\cmidrule{4-7}
& & & TST \cite{zerveas2021transformer}, STraTS \cite{STraTS} & Pre-trains Transformers by predicting masked signal parts & Benefits from parallel processing of Transformer architecture & Focused on specific model structures \\
\cmidrule{4-7}
& & & \cite{ecg_ssl},\cite{har_ssl} & Trains models to recognize data augmentations & Exploits augmentation diversity to find best performing ones & Depends on augmentation relevance \\
\cmidrule{2-7}

& \multirow{25}{*}{\rotatebox{90}{Contrastive-based Methods}} & \multirow{4}{*}{\rotatebox{90}{Inter-sample}} & \cite{simclr}, \cite{simclr_SSC1}, \cite{POPPELBAUM2022_FDsimclr} & Applies SimCLR to time series & Enhances inter-sample learning & Requires careful pair selection \\
\cmidrule{4-7}
& & & MCL \cite{WICKSTROM2022mixupContrastive} & Uses mixup operation for contrastive loss & Novel mixup application & Mixup parameters are critical \\
\cmidrule{4-7}
& & & \cite{triplet_loss} & Searches for positive and negative pairs in input and embedding spaces & Promotes causal learning & Complex pair selection process \\
\cmidrule{3-7}

& & \multirow{15}{*}{\rotatebox{90}{Intra-sample}} & CPC \cite{cpc} & Learns temporal relations by predicting future from past parts in embedding space & Facilitates understanding of temporal dynamics & May neglect non-temporal features \\
\cmidrule{4-7}
& & & SleepDPC \cite{9414752} & Uses multiple context vectors for future timestep prediction in sleep staging & Enhances prediction accuracy for specific tasks & Task-specific design limits general applicability \\
\cmidrule{4-7}
& & & TS-TCC \cite{tstcc} & Applies cross-view prediction tasks with strong and weak augmentations & Improves robustness to augmentation variability & Requires careful design for data augmentations \\
\cmidrule{4-7}

& & & ExpCLR \cite{expCLR} & Leverages expert knowledge for contrastive learning focusing on overlooked properties & Utilizes domain-specific insights effectively & May require access to robust expert features \\
\cmidrule{4-7}
& & & MHCCL \cite{MHCCL} & Employs Masked Hierarchical Cluster-wise Contrastive Learning addressing the issue of false negative pairs & Enhances representation quality by refining cluster prototypes & Complexity in implementing hierarchical clustering and masking strategies \\
\cmidrule{4-7}
& & & MICOS \cite{MICOS} & Combines intra- and inter-class SSL for capturing spatio-temporal features & Effectively utilizes available labels for improved learning & Challenge to balance the different components of the loss function \\
\cmidrule{4-7}
& & & ContrNP \cite{neural_process} & Replaces autoregressive model with a neural process for prediction & Offers flexible context-based predictions & Complexity in model training and tuning \\
\cmidrule{3-7}

& & \multirow{6}{*}{\rotatebox{90}{Frequency}} & BTSF \cite{icml2022iterative} & Utilizes iterative bilinear fusion for time and frequency domain features & Enhances feature representation through fusion & May be computationally intensive \\
\cmidrule{4-7}
& & & STFNets \cite{STFNets} & Combines time and frequency domain augmentations for representation learning & Balances temporal and spectral feature learning & Requires careful balance of domain influences \\
 \cmidrule{4-7}
& & & TF-C \cite{zhang2022_tfConsistency} & Enforces consistency between time and frequency domain representations & Promotes comprehensive feature representation & Complexity in achieving domain consistency \\
\hline
\midrule

\multirow[origin=c]{22}{*}{\rotatebox{90}{Semi-supervised Learning}} & \multirow{12}{*}{\rotatebox{90}{Self-training}} & \multirow{6}{*}{\rotatebox{90}{Pseudo-labeling}}  &\cite{sssl}, \cite{success_semi} & Utilizes model predictions to generate pseudo labels for unlabeled data & Exploits unlabeled data to improve learning & Risk of reinforcing incorrect labels \\
\cmidrule{4-7}
 & & & \cite{fixmatch}, \cite{same_fixmatch}, \cite{selfMatch} & Applies weak and strong augmentations to generate pseudo labels, leveraging self-distillation & Enhances label efficiency and model robustness & Dependent on thresholding and augmentation quality \\
\cmidrule{3-7}

 & & \multirow{5}{*}{\rotatebox{90}{Self-supervised}} & \cite{self_for_semi}, \cite{emadeldeen2022catcc} & Leverages forecasting and fine-tuning on labeled data for semi-supervised learning & Provides robust feature representations & May require extensive pretraining \\
\cmidrule{4-7}
 & & &  \cite{itimes},  \cite{semiTime}, \cite{MtCLSS}, \cite{semiconvnet} & Uses contrastive learning and transformation prediction for semi-supervised learning from time series data & Improves learning from unlabeled data & Complexity in implementation and tuning \\
\cmidrule{2-7}

& \multirow{10}{*}{\rotatebox{90}{Generic Regularization}} & \multirow{1}{*}{\rotatebox{90}{{\scriptsize Ensemble}}} & \cite{teblstm}, \cite{tnnls_semi_sup_work} & Employs temporal ensembling and multi-modal classifiers for leveraging unlabeled data in learning & Reduces the dependency on large labeled datasets & May introduce ensemble disagreement \\
\cmidrule{3-7}
& & \multirow{6}{*}{\rotatebox{90}{Reconstruction}} & HCAE \cite{HCAE}, AnoVAE \cite{ae_seizure}, SMATE \cite{smate} & Trains models to reconstruct input data, leveraging autoencoders and variational autoencoders for semi-supervised learning & Encourages learning of useful representations & Reconstruction quality directly impacts performance \\
\cmidrule{4-7}
 & & & REG-GAN \cite{gan_regression}, SRGAN \cite{gan_fd_pred} & Utilizes GANs for signal regeneration and learning, applying supervised regression on labeled parts & Facilitates learning from unlabeled data & GAN training stability issues \\

\bottomrule
\end{NiceTabular}
\label{table:in_domain_methods}
\end{table*}

\paragraph{Self-supervised Learning}
Instead of pseudo-labeling the unlabeled portion of the data, we can leverage self-supervised learning to learn their representations, and then fine-tune the pretrained model with the available labeled portion. For example, \citet{self_for_semi} propose a forecasting task on the unlabeled data to provide a substitute supervision signal. This task has been jointly optimized along with the supervised task on the labeled part of the data. Also, CA-TCC extends TS-TCC in semi-supervised settings by fine-tuning the pretrained TS-TCC model with the labeled samples \cite{emadeldeen2022catcc}. This fine-tuned model is then used to generate pseudo labels of the unlabeled sample set, which are then deployed to train the model with a supervised contrastive loss.
In addition, iTimes~\cite{itimes} learns from unlabeled samples by applying several transformations to the time series via irregular time sampling techniques and learns the model to predict the transformation type to learn the temporal structure.

Contrastive learning has also been widely used for self-supervised pretraining in semi-supervised learning. For example, SemiTime uses contrastive learning by splitting the signal into two parts, i.e., past and future parts \cite{semiTime}. It forms positive pairs among the past and the future parts of the same signal, and negative pairs between the past of the signal and the future of other signals. Moreover, MtCLSS applies contrastive learning between the feature representations of the labeled and unlabeled EEG samples for the sleep stage classification problem \cite{MtCLSS}. TS-TFC~\cite{Liu_Ma_Ma_Wang_2023} leverages time- and frequency-domain views to dual deep neural networks and uses the supervised contrastive learning module to adjust for category-specific learning difficulties. Through cross-view pseudo-labeling and co-training, TS-TFC effectively captures complementary information from both domains. Last, Semi-DeepConvNet uses a SimCLR-like contrastive learning technique to learn from EEG-based motor imagery data \cite{semiconvnet}. It also eliminates the bias to the data from the different subjects by using an adversarial training scheme.

An orthogonal approach was proposed by TC-SSL~\cite{TC-SSL}, which presents a strategy to improve model accuracy and efficiency by focusing on time consistency in unlabeled data selection. The idea is to identify and prioritize unlabeled samples with consistent model outputs throughout training, while incorporating a dual self-supervised loss—consistency between a sample and its augmentation, alongside a contrastive loss for distinct sample outputs.

\subsubsection{Generic regularization}
Apart from self-training, another stream of methods adopts different techniques to learn from data, e.g., ensembling and generative models. 

\paragraph{Ensemble-based Methods}
These methods train two or more models separately on the same dataset or on different views of the data.
For example, TEBLSTM proposes temporal ensembling for human activity recognition data based on an LSTM model \cite{teblstm}. In addition to the supervised cross-entropy loss on the labeled samples, they train the unlabeled data by comparing their predictions with the ensembled predictions of the past epochs. Moreover, \cite{tnnls_semi_sup_work} cluster labeled samples using k-means to train multi-modal classifiers. If the ensemble of these classifiers reaches an agreement on predicting an unlabeled sample, it is assigned a label and added to the labeled set. The newly labeled samples are then added to the training set and the models are retrained on the expanded dataset.

\paragraph{Reconstruction-based Methods}
These methods train the model to reconstruct input data from a latent representation assuming that the model will learn useful representations from this process. For example, HCAE trains an autoencoder on industrial motor bearing datasets to reconstruct the spectrogram of the signals \cite{HCAE}. They train a supervised loss on the labeled samples besides the reconstruction loss. Similarly, AnoVAE trains a variational autoencoder on normal EEG signals and marks the anomalous signals by the trained model as seizures \cite{ae_seizure}. Also, SMATE employs a reconstruction loss to learn from data, uses the labeled samples to create centroids, and then classifies the unlabeled data according to their distance to the centroids \cite{smate}.

Moreover, other methods use GANs to reconstruct and learn from data. For example, REG-GAN is a semi-supervised regression method that uses GAN to learn representations by regenerating the signals of the whole data and applying supervised regression loss on the labeled part \cite{gan_regression}. SRGAN proposes a GAN model to learn useful information from suspension histories to improve the remaining useful life prediction performance of industrial machines \cite{gan_fd_pred}.

\section{Cross-domain Learning}
Cross-domain learning refers to training a model on a labeled domain (i.e., the source domain), and transferring the knowledge to a related domain (i.e., the target domain), which is the domain of interest. This can be a challenging problem, as we may need to train the model to adapt the different distributions across the two domains.

\begin{figure}
    \centering
    \includegraphics[width=\columnwidth]{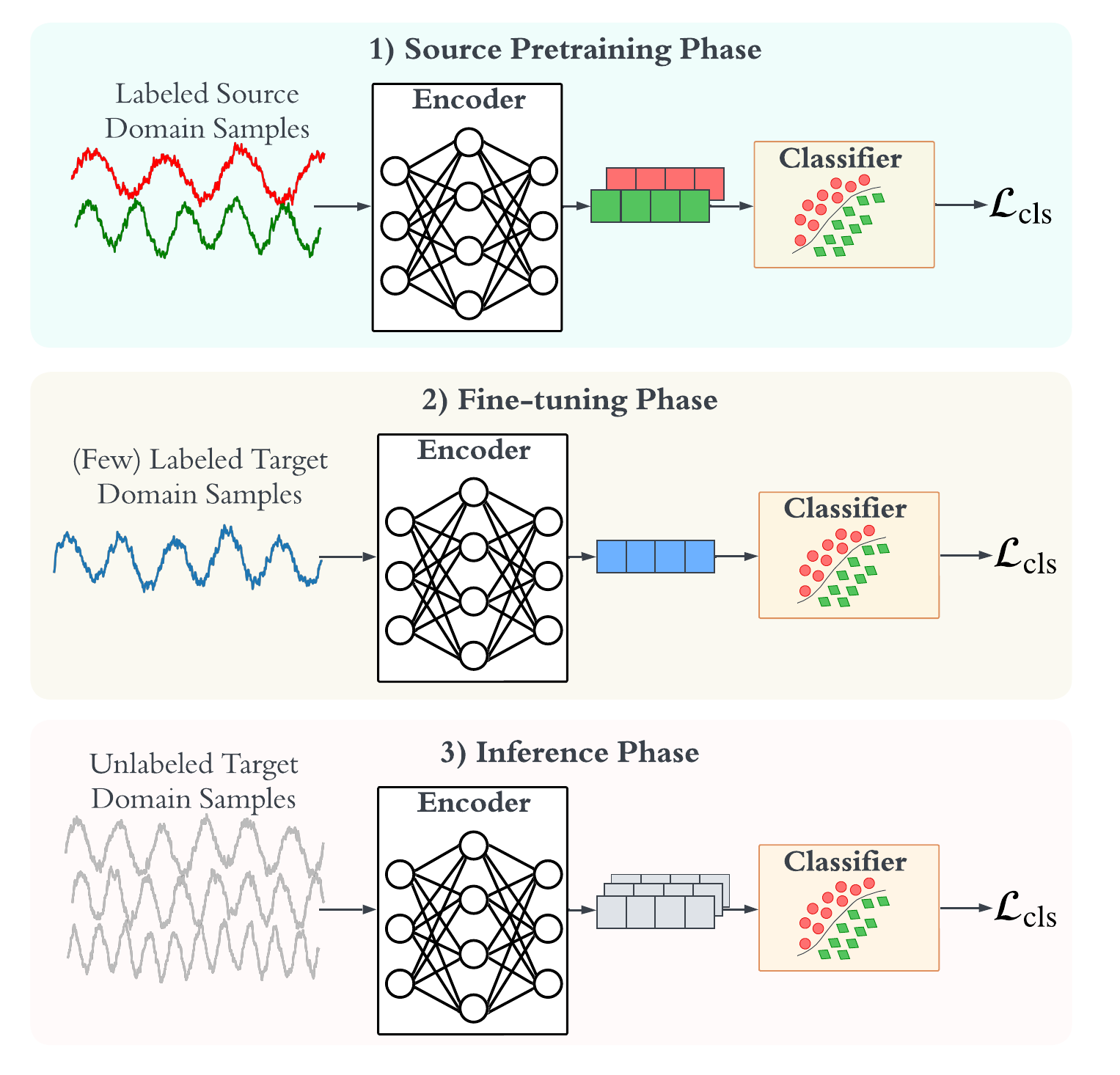}
    \caption{The three phases of the transfer learning process. In the first phase, the model is trained on the labeled source domain data. In the second phase, the pretrained model is fine-tuned on the available few target domain samples. Last, the model is tested on the unlabeled target domain data during the inference phase.}
    \label{fig:transfer_learning}
\end{figure}

\subsection{Transfer Learning}
Transfer learning is used when only a few labeled data are available in the target domain, while more labeled samples are provided in the source domain. With a source domain dataset \( \mathcal{D}_S = \{(x^S_i, y^S_i)\} \) and a target domain dataset \( \mathcal{D}_T = \{(x^T_j, y^T_j)\}_{j=1}^{M} \), transfer learning seeks to transfer knowledge from \( \mathcal{D}_S \) to \( \mathcal{D}_T \). This involves training a model \( f \) on \( \mathcal{D}_S \) and then fine-tuning, adapting, or directly applying it to \( \mathcal{D}_T \).

Thus, the model can be pretrained on the source domain data, and then directly fine-tuned with the few labeled samples in the target domain, as illustrated in Fig.~\ref{fig:transfer_learning}.
\citet{NIPS2014_375c7134} perform one of the first works to study the impact of transfer learning, where they found that the features produced by the first layer in deep neural networks tend to be general and can be applied to many datasets and tasks.
The idea behind transfer learning is straightforward, yet found to be effective, and has been directly applied in different time series applications.
For example, \citet{tl_mi} demonstrate the positive impact of transfer learning to improve the performance of EEG-based stroke rehabilitation for an effective motor imagery system. 
Also, \citet{ZHONG2022765} conclude that transfer learning helps reduce the model parameters while increasing its accuracy in the fault diagnosis problem. They propose SLTL, which combines a self-attention mechanism with transfer learning within a lightweight CNN model. This model transforms vibration signals into time-frequency images and applies an optimized SqueezeNet model.

Transfer learning can be particularly impactful in time series applications that include multiple users/subjects, such as healthcare or human activity recognition. In such cases, we can improve the model performance on new unseen users with transfer learning instead of training the model from scratch \cite{tl_ts_subjects}.
For example, \citet{TL_sensor_clf} propose to train a single-channel neural network using single-channel data from the source domain on human activity recognition datasets. Then, they replicate these single-channel nets for each target domain channel, showing improved performance.

It is worth noting that the positive impact of transfer learning is not guaranteed in all cases. It has been found that the effect of transfer learning depends on the relationship between the source and target distributions. To demonstrate this phenomenon, \citet{fawaz_tl} evaluate the impact of transfer learning on 85 UCR datasets by training on one dataset and testing on all the others. They find that transfer learning does not guarantee a positive effect on all the cross-dataset scenarios. Therefore, they propose DTW as a measure to predict the effectiveness of transfer learning given a source and target pair. Following up, \citet{tl_select_source} also use DTW and Jensen-Shannon divergence to select the suitable source domain depending on the inner distribution dissimilarity to the target domain. Similarly, \citet{tl_rank_src_dom} rank the multiple source domains according to the Mean Silhouette Coefficient, which measures how well each source domain can cluster the target domain in the embedding space.

\subsection{Unsupervised Domain Adaptation}
\begin{figure}
    \centering
    \includegraphics[width=\columnwidth]{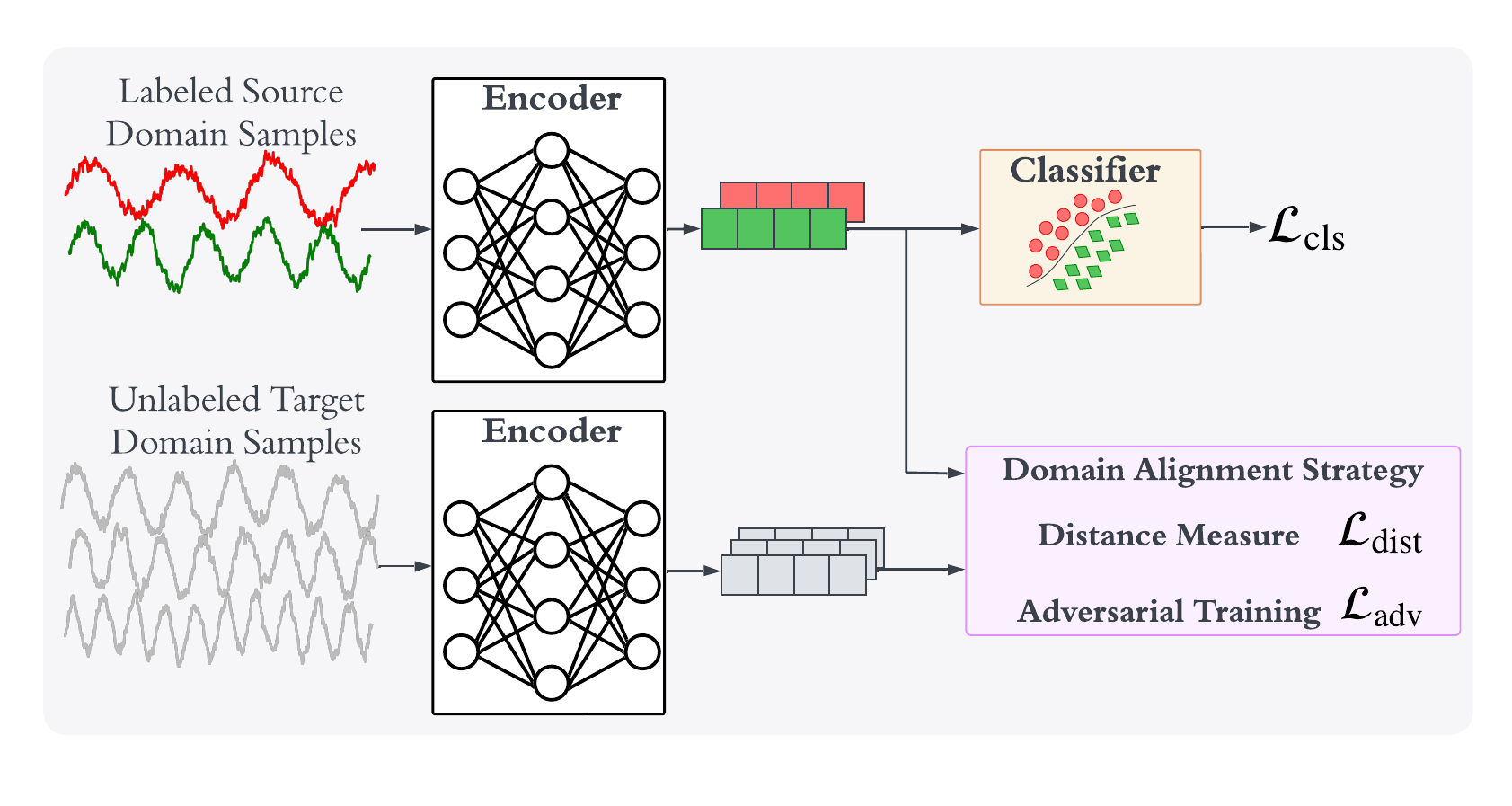}
    \caption{A description of the unsupervised domain adaptation process. The model is fed with both the labeled source domain data and the unlabeled target domain data. The model is trained to reduce the domain shift between the two domains with an adaptation technique, such as reducing a distance measure or applying an adversarial training strategy.}
    \label{fig:domain_adaptation}
\end{figure}

Unsupervised domain adaptation (UDA) is a subset of transfer learning, that is required when no target domain labels are available and when a distribution shift exists between source and target domains. UDA adapts a model trained on a labeled source domain to a shifted unlabeled, but related, target domain, to minimize the domain shift. 
Given a source domain dataset \( \mathcal{D}_S = \{(x^S_i, y^S_i)\}_{i=1}^{N_S} \) with \( N_S \) labeled instances, and a target domain dataset \( \mathcal{D}_T = \{x^T_j\}_{j=1}^{N_T} \) with \( N_T \) unlabeled instances, the objective is to learn a predictive model \( f \) that performs well on \( \mathcal{D}_T \), leveraging the labeled data from \( \mathcal{D}_S \) and the unlabeled data from \( \mathcal{D}_T \).
In general, methods proposed for time series UDA can be categorized into distance-based and adversarial-based methods. The training process of UDA is illustrated in Fig.~\ref{fig:domain_adaptation}. 

\subsubsection{Distance-based Methods}
The main idea behind distance-based UDA is to minimize a distance measure between the source and target distributions so that the model can learn to generalize to the target domain.
There are several distance metrics, e.g., Maximum Mean Discrepancy (MMD) \cite{mmd} and Correlation Alignment (CORAL) \cite{coral}, that have been used in the visual UDA works. 
For example, the MMD loss is a non-parametric measure of the distance between the source and target domain distributions:
\begin{equation}
    \mathcal{L}_{MMD} = \left\|\frac{1}{N_S}\sum_{i=1}^{N_S}\phi(g(x^S_i)) - \frac{1}{N_T}\sum_{j=1}^{N_T}\phi(g(x^T_j))\right\|^2,
\end{equation}
where \(\phi\) represents the mapping of features to a high-dimensional space, aiming to minimize the difference in distributions of the source and target domains in this space.
However, since these distance measures are found to be less effective in mitigating the large distribution shift, some works modified these measures to achieve better performance.
For example, AdvSKM refines the MMD measure to be suitable for time series, by updating the inner kernel of MMD to become a hybrid spectral kernel network \cite{advskm}.
In addition, \citet{ts_cov_shift} pretrain the model on the unlabeled target domain data, then train an optimal transport jointly with CORAL to transform the source to the target domain. They also measure the class-wise similarity between the source and target embeddings to select the best transformation and use it to classify transformed embedding with the target domain model.

Also, SASA considers the variant information in time series and learns the sparse associative structure by studying the causal structure of time series variables, building on top of an LSTM architecture \cite{sasa_ts}.
Further, \citet{SPS} proposed soft parameter sharing (SPS) domain adaptation architecture for time series, overcoming the limitations of traditional hard parameter sharing methods. They enhance adaptation by employing a non-linear relation between source and target models and refining the adaptation loss with squared MMD.
Additionally, RAINCOAT~\cite{RAINCOAT} introduces an approach to both closed-set and universal domain adaptation for complex time series, effectively managing feature and label shifts through temporal and frequency feature alignment and misalignment corrections. This model recognizes domain-specific labels and identifies label shifts, enhancing transfer learning across domains. 
Differently, MAD~\cite{MAD} harmonizes feature and temporal shifts between source and target domains. By leveraging Optimal Transport (OT) loss and DTW within a deep learning framework, MAD aligns time series data and enhances discriminative power.

The distance-based UDA methods are usually simpler in implementation and faster in training. However, they may not be as effective as adversarial-based methods in reducing the large domain shift between source and target domains. In addition, they may not efficiently capture the underlying structure of the data as well as adversarial methods.

\subsubsection{Adversarial-based Methods}
Adversarial-based UDA bridges the gap between source and target domains with generative adversarial training \cite{adda}. In specific, the source and target data are passed through a discriminator network, which is trained to make the two domains indistinguishable. 
Formally, the model \( f \) comprises a feature extractor \( g \) and a classifier \( h \), with an additional domain discriminator \( d \) introduced to differentiate between source and target domain features. The domain adversarial loss is formulated as:
\begin{equation}
\mathcal{L}_{adv} = -\sum_{i=1}^{N_S} \log(d(g(x^S_i))) - \sum_{j=1}^{N_T} \log(1 - d(g(x^T_j))).
\end{equation}
This loss encourages the feature extractor \( g \) to generate domain-invariant features such that the domain discriminator \( d \) cannot distinguish between the source and target domains. Notably, the performance of this loss function is dependent on the architecture of the domain discriminator \( d \) and the feature extractor \( g \). Proper tuning of these components is crucial for balancing discrimination and generalization.
The common architecture for this discriminator is either a fully connected or a CNN-based network. However, SLARDA proposes an autoregressive domain discriminator that considers the temporal features when distinguishing source and target domains \cite{slarda}. It also pretrains the encoder with a forecasting task to boost the performance. Finally, a mean-teacher model is used to learn the fine-grained class distribution of the target domain.

In most works, the feature extractor architecture is usually shared between source and target domains. However, Domain Adaptation Forecaster (DAF) uses an unshared feature extractor and a shared attention-based architecture for time series UDA forecasting \cite{daf_icml}. The shared attention module keeps the domain-specific features, and the domain discriminator leverages the attention output to perform the adversarial training. Similarly, ADAST proposes a shared feature extractor and an unshared domain-specific attention model \cite{adast}. Also, ContrasGAN leverages a bidirectional GAN to adapt the source and target domains with two unshared generators and discriminators to regenerate samples between domains \cite{ContrasGAN}. They also discriminate classes by leveraging contrastive learning with adaptation to minimize the intra-class discrepancy. Also, \citet{XI2023147} propose an unsupervised multimodal domain adversarial network for time series classification. They employ dual feature extractors for time-domain and frequency-domain representations, fused via a domain discriminator and a Time-Frequency-domain joint MMD approach for effective source-target distribution alignment.

To benefit from the availability of multiple source domains in human activity recognition, CoDATS proposes a weakly supervised multi-source UDA \cite{codats}. It learns domain invariant features with adversarial training, based on a gradient reversal layer (GRL) to flip the gradients when being back-propagated.

Adversarial-based UDA methods are computationally expensive and can be more difficult to implement than distance-based UDA methods. However, they are more effective at reducing the large domain shift between source and target domains.

Another line of work attempts to address the UDA problem from a different perspective. For example, COTMix \cite{cotmix} proposes a contrastive-based approach to mitigate the domain shift. The authors attempt to move the source and target domains to an intermediate domain with a contrastive loss.
Furthermore, \citet{MAPU} present an approach to adapt the unlabeled target domain to a labeled source domain in a source-free manner, i.e., the adaptation occurs without accessing the source domain data. Such a method promotes data privacy.
Also, SEA~\cite{SEA} addresses domain adaptation by aligning features at both local sensor-specific features and correlations and global overall sensor features levels, while a multi-branch self-attention mechanism captures spatial-temporal dependencies using graph neural networks. However, this method can be only applied to the multivariate time series.

\subsection{Semi-supervised Domain Adaptation}
\begin{figure}
    \centering
    \includegraphics[width=\columnwidth]{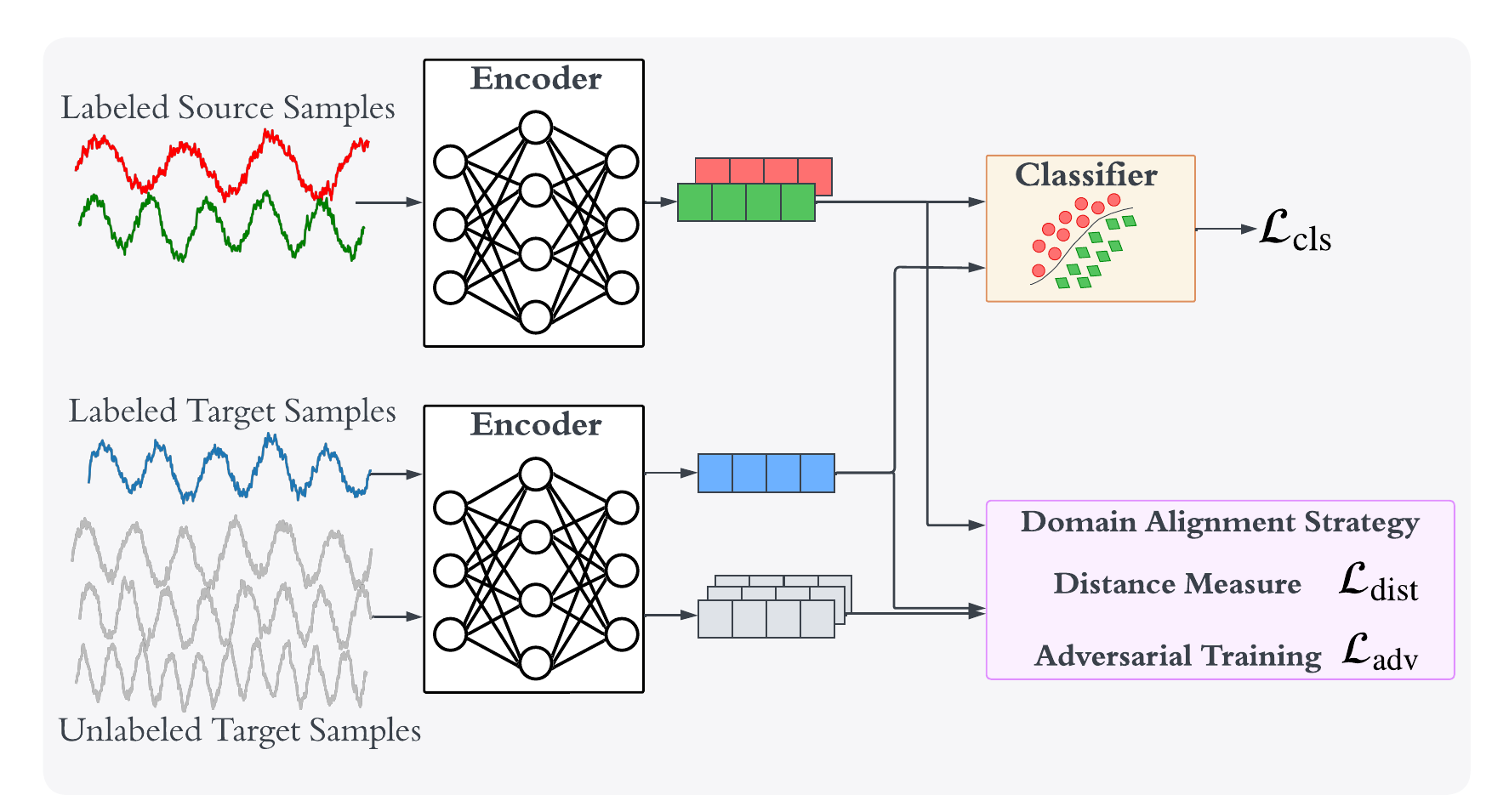}
    \caption{The semi-supervised domain adaptation process. The process is similar to the unsupervised domain adaptation with a classification loss applied to the labeled samples from both the source and target domains.}
    \label{fig:semi_domain_adaptation}
\end{figure}

Semi-supervised domain adaptation is used to adapt a model trained on a source domain to a related out-of-distribution target domain, in which only a small amount of labeled data is available for the target domain.
In addition to the source domain dataset \( \mathcal{D}_S \) and the target domain dataset \( \mathcal{D}_T \), assume a subset of \( \mathcal{D}_T \) is labeled: \( \mathcal{D}_{T_L} = \{(x^T_k, y^T_k)\}_{k=1}^{N_{T_L}} \). The objective is to leverage \( \mathcal{D}_S \), \( \mathcal{D}_{T_L} \), and the unlabeled portion of \( \mathcal{D}_T \) to learn a model \( f \) that achieves high performance on the target domain.
The labeled samples in the target domain can be helpful in two aspects. First, it can be used to select the best target model hyperparameters. Second, these samples can be used to boost the adaptation performance of the model. The training process of the semi-supervised domain adaptation is shown in Fig.~\ref{fig:semi_domain_adaptation}, where we apply the classification loss to the labeled samples from both the source and target domains and use a domain alignment strategy to align the distribution of both domains.

Most previous semi-supervised UDA works have adopted the adversarial-based approach.
For example, \citet{heremans2022unsupervised} generate pseudo labels to boost the performance along with the target domain labeled data while performing adversarial domain adaptation for EEG signals.
Also, \citet{semi_uda_fd} propose nuclear norm maximization and consistency loss with virtual adversarial training to measure the discrepancy between the predictions of the unlabeled samples and their perturbations, which helps to maintain diversity and discriminability. Those components are added along with the adversarial training on fault diagnosis data.

Semi-supervised UDA is less explored despite being more practical in real-world scenarios, probably due to the increased challenge of labeling the target domain samples.

\begin{table*}[!h]
\centering
\caption{Summary of Cross-domain Methods for Time Series Representation Learning. For space limitations, Pseudo Labeling is abbreviated as `P-L', and Consistency loss is abbreviated as `Consis.'}
\begin{NiceTabular}{c|c|p{1.7cm}|p{4.9cm}|p{4cm}|p{4cm}}
\toprule
\multicolumn{2}{c|}{\textbf{Category}} & \textbf{Method} & \textbf{Description} & \textbf{Advantages} & \textbf{Potential Limitations} \\

\midrule
\multirow[origin=c]{8}{*}{\rotatebox{90}{Transfer Learning}} & & \cite{NIPS2014_375c7134} & Investigates the impact of transfer learning, finding first-layer features in deep neural networks to be general across tasks & Demonstrates the general applicability of learned features & May not address task-specific nuances \\
\cmidrule{3-6}
~ & &  \cite{tl_mi} & Uses transfer learning to improve EEG-based MI systems for stroke rehabilitation & Enhances performance with limited labeled data & Requires careful adaptation to target domain \\
\cmidrule{3-6}
~ &  &SLTL \cite{ZHONG2022765} & Combines CNN and self-attention with transfer learning for fault diagnosis & Reduces model parameters while increasing accuracy & Dependent on the quality of time-frequency image transformation \\
\hline
\midrule


\multirow[origin=c]{33}{*}{\rotatebox{90}{Unsupervised Domain Adaptation}} & \multirow{13}{*}{\rotatebox{90}{Distance-based}} & AdvSKM \cite{advskm} & Refines MMD for time series with a hybrid spectral kernel network & Tailors distance measures for time series specificity & May require careful kernel design \\
\cmidrule{3-6}
& &  \cite{ts_cov_shift} & Combines optimal transport and CORAL for domain transformation & Enhances source to target domain alignment & Complexity in optimal transport computation \\
\cmidrule{3-6}
&  & SASA \cite{sasa_ts} & Learns sparse associative structures within time series data using LSTM & Captures causal structures effectively & Limited by LSTM's handling of long sequences \\
\cmidrule{3-6}
&  & SPS \cite{SPS} & Implements soft parameter sharing for domain adaptation & Offers flexible adaptation strategy & Non-linear relation management can be challenging \\
\cmidrule{3-6}
&  & RAINCOAT \cite{RAINCOAT} & Manages feature and label shifts with temporal and frequency alignment & Addresses both closed-set and universal adaptation challenges & May involve intricate alignment processes \\
\cmidrule{3-6}
&  & MAD \cite{MAD} & Utilizes Optimal Transport and DTW for domain alignment & Enhances discriminative power of time series data & Integrating OT and DTW can be computationally demanding \\
\cmidrule{2-6}

& \multirow{12}{*}{\rotatebox{90}{Adversarial-based}} & SLARDA \cite{slarda} & Employs autoregressive discriminator, SSL, and a mean-teacher model & Leverages temporal dynamics for domain distinction & Requires pretraining for optimal performance \\
\cmidrule{3-6}
&  & DAF \cite{daf_icml} & Uses unshared feature extractors with shared attention for time series forecasting & Keeps domain-specific features intact & Complexity in attention mechanism adaptation \\
\cmidrule{3-6}
&  & ADAST \cite{adast} & Combines shared feature extraction with unshared domain-specific attention models & Tailors domain adaptation with attention detail & Balancing shared and unshared components is critical \\
\cmidrule{3-6}
&  & ContrasGAN \cite{ContrasGAN} & Uses bi-GAN and contrastive learning for intra-class discrepancy minimization & Effectively regenerates samples between domains & Complexity in GAN training and contrastive loss management \\
\cmidrule{3-6}
&  & CoDATS \cite{codats} & Weakly supervised multi-source UDA with adversarial training and GRL & Enhances domain invariant feature learning & Multi-source domain integration can be complex \\
\cmidrule{2-6}

& \multirow{8}{*}{\rotatebox{90}{Other Approaches}} & COTMix \cite{cotmix} & A contrastive-based approach to mitigate domain shift & Uses lightweight contrastive loss for domain adaptation & Works on domains having similar timestep length \\
\cmidrule{3-6}
& & MAPU \cite{MAPU} & Adapts unlabeled target domain in a source-free manner, promoting data privacy & Addresses adaptation without accessing source data & May face challenges in capturing complex domain shifts \\
\cmidrule{3-6}
& & SEA \cite{SEA} & Aligns local and global features with a multi-branch self-attention using GNNs & Captures spatial-temporal dependencies effectively & Requires sophisticated model architecture design \\
\hline
\midrule

\multirow[origin=c]{6}{*}{\rotatebox{90}{Semi-Sup. DA}} & \multirow{3}{*}{\rotatebox{90}{P-L}} &  \cite{heremans2022unsupervised} & Utilizes adversarial domain adaptation for EEG signals, enhancing performance with pseudo labels and labeled target data & Improves adaptation by incorporating target domain supervision & Relies on the quality of pseudo labels \\
\cmidrule{2-6}
& \multirow{3}{*}{\rotatebox{90}{Consis.}} &  \cite{semi_uda_fd} & Combines nuclear norm maximization with adversarial training and consistency loss for fault diagnosis & Ensures robust adaptation by promoting feature diversity & Complexity in balancing loss components \\
\bottomrule

\end{NiceTabular}
\label{table:cross_domain_methods}
\end{table*}

\section{Discussion and Future Directions}
In this section, we outline the pros and cons of the above categories, as well as the potential future research directions that stem from the insights, garnered through our review of label-efficient techniques for time series representation learning.

\subsection{Incorporating Hybrid Approaches}
One intriguing direction lies in the fusion of various label-efficient techniques, which has the potential to harness the complementary strengths of each approach. This concept stems from the recognition that no single technique can address all the challenges associated with time series analysis, especially when dealing with its diverse and complex structure. For example, combining data augmentation, self-supervised learning, and semi-supervised learning within a unified framework has the potential to harness the complementary strengths of each approach. Such hybrid strategies could enable models to leverage both the rich information encoded in the raw data and the intrinsic relationships within the labeled samples. 

Exploring the synergies between these techniques and assessing their combined impact on performance and efficiency presents an exciting area of research. This has been studied in \cite{aug_tl}, where the authors propose generating synthetic time series data with diversified characteristics from the original dataset and using that as a source dataset for transfer learning tasks. Also, \citet{dasting} generate augmented data to increase the patterns variability in the dataset, and then deploy domain adversarial transfer learning for time series forecasting.
Besides augmentations, self-supervised learning has also shown a high learning capability from unlabeled data, which promotes it for learning from unlabeled data in any scenario. Hence, numerous semi-supervised learning and transfer learning-based methods integrate self-supervised learning in their methodologies, as in \cite{emadeldeen2022catcc,ContrasGAN}. 

Notably, this fusion approach can raise some challenges, e.g., the increased model complexity, which requires careful design, tuning, and validation to avoid overfitting or other unintended consequences. In addition, the combination of different techniques might impose higher computational requirements, necessitating optimization both in algorithmic design and computational resources. Last, it is important to investigate how can we tailor the fused models to specific domains, such as finance or healthcare, to increase its effectiveness in addressing domain-specific challenges.

\subsection{In-Domain vs. Cross-Domain Solutions: A Strategic Decision}
The landscape of label-efficient time series representation learning offers both in-domain and cross-domain solutions, each with unique characteristics and applications. In deciding whether to rely on in-domain or cross-domain solutions, it is important to consider the context, goals, and challenges specific to the data.
This section explores the factors that guide the selection between these two paradigms, the potential synergies, and the areas where future research is needed.

On one hand, the in-domain solutions are usually tailored to the characteristics of the specific time series domain, allowing them to capture nuances, patterns, and context unique to that domain. This can lead to more accurate and relevant representations. In addition, it helps the preservation of the domain structure. Techniques like data augmentation and self-supervised learning can preserve the inherent structure of the data, enabling models to learn from perturbed versions of real-world instances.
However, in-domain techniques may struggle to generalize to unseen domains or scenarios, especially when the labeled data is scarce or specific to a narrow context. On the other hand, cross-domain solutions capitalize on knowledge transfer from related domains, which makes models robust across different domains, enabling them to perform well on new and unseen data distributions. However, the success of cross-domain solutions heavily depends on the degree of similarity between the source and target domains. Specifically, we may suffer from the risk of negative transfer if domains are not sufficiently aligned, besides the increased computational complexity in matching and adapting across domains.

Therefore, the decision between in-domain and cross-domain solutions is not a binary one, but rather a strategic choice shaped by the nature of the task, the availability of data, and the need for generalizability. Future research that explores the synergies between these paradigms, the development of decision-making frameworks, and the ethical and practical implications will contribute to a more nuanced understanding of label-efficient time series representation learning.

\subsection{Choosing the Best Method within Each Category}
Selecting the most suitable label-efficient method within each category, e.g., self-supervised learning, presents a challenge due to the diversity of techniques available. The decision should be guided by the specific context of the problem, the available resources, and the desired outcome. For instance, assume a scenario where we have unlabeled data and we decided to use contrastive self-supervised learning. It might be confusing to choose between inter-sample or intra-sample contrasting methods. This could be attributed to the lack of consistency in the backbone networks, the evaluation schemes, and the evaluating datasets among existing methods. Therefore, it can be challenging to identify the best-performing methods under each category \cite{emadeldeen2023eval}. 

The creation of benchmarks is a forward step into addressing this challenge. Benchmarks provide standardized evaluation platforms that enable fair comparisons between different methods. In addition, they facilitate insightful comparisons of strengths, weaknesses, and limitations across different techniques. Moreover, they help researchers identify key challenges and gaps, driving the development of new methods that address specific issues highlighted by benchmark performance.
One example of these benchmarks is AdaTime \cite{adatime}, which provides a unified benchmarking suite to the UDA algorithms on time series data. Therefore, the development of dedicated benchmarking suites is a promising future direction.

\subsection{Weakly Supervised Learning}
Weakly Supervised Learning (WSL) represents a pivotal shift in deep learning paradigms, especially for time series data, by training models on indirect or noisy labels rather than relying on extensively labeled datasets \cite{wsl}. The essence of WSL lies in its ability to significantly reduce the labeling effort—a critical advantage over traditional fully supervised methods, which demand point-wise labels for every data instance. This reduction is particularly beneficial in scenarios with limited data availability, enabling models to extrapolate meaningful information from sparse datasets. It also facilitates the integration of domain-specific knowledge directly into the learning process, through crafted labeling functions or constraints, enhancing model relevance and accuracy \cite{10292790}. 

This approach is built upon various forms, such as \textit{partial labels}, where only a subset of the data points within a time series is labeled. Notable contributions in this area include the work of \citet{10.1007/978-3-319-68765-0_17}, that introduced a method for learning discriminative patterns, or shapelets, from partially labeled time series data, enhancing tasks such as classification and clustering. \citet{10.5555/3454287.3454305} applied WSL to the classification of freezing episodes in Parkinson's patients using data from wearable sensors and heuristic labeling functions.
Another form of WSL can be built upon \emph{relational labels}, which involves labels that describe relationships between time series segments, such as similarity or sequential order. For example, \citet{MENATORRES20144224} explored classification based on pairwise similarities, developing a model that learns a distance metric to capture these relationships.

\subsection{Active Learning}
Active learning allows the model to interact with its environment and actively seek labels on data that it is uncertain about, instead of relying solely on passively provided labeled data \cite{settles2009active}. This process often involves human intervention, where the model identifies and requests labels for the most informative data points from human annotators during training. In this way, active learning enables the model to selectively focus on acquiring labels that are likely to improve its performance and reduce the overall amount of labeled data needed.

Despite its potential, few works have been proposed for active learning on time series data. For example, \citet{active_semi_ts} propose a semi-supervised active learning approach based on human activity recognition. They rely on self-training to select the most relevant samples for annotation and training the model.
Also, the Transfer Active Learning (TAL) method selects the candidate samples by evaluating their informativeness and representativeness \cite{TAL}. These criteria are calculated by mapping the input sample into the embedding space from both sample and sample-label views.

Designing task-specific active learning techniques for time series data, exploring synergies with self-supervised learning, and establishing evaluation protocols that reflect real-world scenarios can be promising future directions in this field.

\subsection{Prototypical networks}
Prototypical Networks is another approach for addressing the label scarcity issue in various machine learning applications, including time series representation learning. These networks build on the idea of learning a metric space where representations of examples are computed to reflect the underlying class structure \cite{NIPS2017_cb8da676}. Prototypical Networks rely on the concept of learning a prototypical representation for each class, derived from the embeddings of examples within that class. Prototypes are then computed as the mean of the embeddings of the examples within a class. After that, unlabeled instances are compared to these prototypes using a distance metric, e.g., Euclidean distance, and classification is performed based on proximity to the prototypes.

With this design, Prototypical Networks can inherently operate effectively with few labeled examples, making them an appealing solution for label scarcity. Despite that, few works have adopted this approach. For example, \citet{zhang2020tapnet} propose TapNet as an attentional prototype network for multivariate time series data. Despite the potential of prototypical networks, challenges such as handling long sequences, addressing high-dimensional data, and mitigating the impact of noise must be addressed.

\section{Conclusion}
The scarcity of annotated data presents a significant barrier to the effective deployment of deep learning models in real-world scenarios. This paper aimed to offer a holistic perspective, seamlessly connecting disparate research works and carving out a clear roadmap for tackling various label-deficient situations and their potential remedies. Our discourse underscores the importance of selecting the appropriate deep learning architecture, and the imperative to understand the nature of available time series data when determining the suitability of in-domain versus cross-domain approaches. Our investigation further highlights the potential of fusing multiple techniques to yield a better performance. Moreover, this survey underscores the necessity for standardized benchmarks, which would streamline the decision-making process when navigating the plethora of methodologies within each category.

\bibliographystyle{IEEEtranN}
\bibliography{references.bib}

\end{document}